\pdfoutput=1

\documentclass[runningheads]{llncs}
\usepackage[T1]{fontenc}

\input{preamble}

\usepackage{graphicx,verbatim}
\usepackage{booktabs}
\usepackage{amsfonts}
\usepackage{amsmath}
\usepackage{bm}
\usepackage{booktabs}
\usepackage{makecell}
\usepackage{caption}
\usepackage{subcaption}
\usepackage{hyperref}
\usepackage{pifont}
\usepackage{adjustbox}

\begin{document}
\title{Feature Space Guidance for Breast Cancer Classification in DCE-MRI}
\titlerunning{Feature Space Guidance for Breast Cancer Classification}

\author{Benjamin Hamm\inst{1,4} %
\and Yannick Kirchhoff\inst{1,2,3} %
\and Maximilian Rokuss\inst{1,2,3,5} %
\and Moritz Langenberg\inst{1,2} %
\and Constantin Ulrich\inst{1} %
\and Tassilo Wald\inst{1,2,5} %
\and Jeremias Traub\inst{1,5} %
\and Karol Gotkowski\inst{1,5} %
\and Klaus Maier-Hein\inst{1,4,6} %
}
\authorrunning{Hamm et al.}
\institute{German Cancer Research Center (DKFZ) Heidelberg, Division of Medical Image Computing, Germany
\and
Faculty of Mathematics and Computer Science, Heidelberg University, Germany
\and
HIDSS4Health -- Helmholtz Information and Data Science School for Health, Karlsruhe/Heidelberg, Germany
\and
Medical Faculty, Heidelberg University, Germany
\and
Helmholtz Imaging, German Cancer Research Center (DKFZ), Heidelberg, Germany
\and
Pattern Analysis and Learning Group, Department of Radiation Oncology, Heidelberg University Hospital, Germany\\
    \email{benjamin.hamm@dkfz-heidelberg.de}}
  
\maketitle              %
\begin{abstract}

Dynamic contrast enhanced breast MRI (DCE-MRI) is a powerful clinical tool for breast cancer detection, providing high resolution anatomical detail together with rich temporal contrast information. However, high dimensional 4D inputs, small lesions, and heterogeneous acquisition protocols across clinical sites hinder robust automated classification of healthy, benign, and malignant cases. To address these challenges, we propose a framework that dynamically analyzes latent representations to adapt to protocol-specific characteristics. Spatial variability is mitigated by reducing confounding background uptake and compensating for misalignment caused by deformable soft tissue. Additionally, relationships in the latent space across phases are leveraged to select the most informative temporal features, improving robustness to protocol-specific temporal variability. Finally, task specific discriminative features are promoted through large scale supervised lesion segmentation pretraining, which substantially enhances downstream finetuning. Evaluated under leave-one-center-out validation on the ODELIA dataset and the held-out AMBL cohort, the proposed framework substantially outperforms finetuned radiology foundation models and prior methods, improving mean AUROC by nearly 8 points and balanced accuracy by 4 points over the strongest baseline. Additionally, our method achieved first place in the MICCAI ODELIA Breast MRI Challenge 2025, further demonstrating its effectiveness for robust breast cancer classification. We publicly release our codebase under \url{https://github.com/MIC-DKFZ/CURIAtor}.

\keywords{Breast Cancer Classification \and Adaptive Temporal Modeling \and Dynamic Contrast Enhanced MRI}

\end{abstract}

\section{Introduction}
Breast cancer remains one of the leading causes of cancer-related mortality among women worldwide \cite{sung2021global}. Early and accurate diagnosis is therefore critical for improving patient outcomes and guiding personalized treatment \cite{saadatmand2015influence}. Among available imaging modalities, dynamic contrast-enhanced magnetic resonance imaging (DCE-MRI) has emerged as a powerful tool for breast cancer assessment, as it provides both high-resolution anatomical information and rich temporal signatures of tissue perfusion and permeability \cite{mann2019breast}. These functional characteristics enable improved sensitivity compared to conventional mammography and ultrasound, particularly in dense breast tissue \cite{riedl2015triple}. Approaching this as a classification of healthy, benign, and malignant cases is attractive because case-level labels are easier to obtain than dense segmentation annotations, making the task formulation well suited for scalable screening \cite{saldanha2025swarm}.

Despite its clinical value, reliable automated classification of DCE-MRI remains a challenging problem. A typical examination consists of large three-dimensional volumes acquired at multiple time points following contrast injection, resulting in high-dimensional, heterogeneous data \cite{kataoka2024ultrafast}. Lesions often occupy only a small fraction of the imaged volume, resulting in extreme foreground sparsity that weakens discriminative signal extraction. Moreover, acquisition protocols vary substantially across scanners and institutions, with differences in temporal resolution, number of phases, and contrast timing. This variability induces distribution shifts that degrade the performance and reliability of supervised learning systems trained on limited or homogeneous datasets \cite{alanazi2026domain}.

\begin{figure}[!t]
 \centering
 \includegraphics[width=\linewidth]{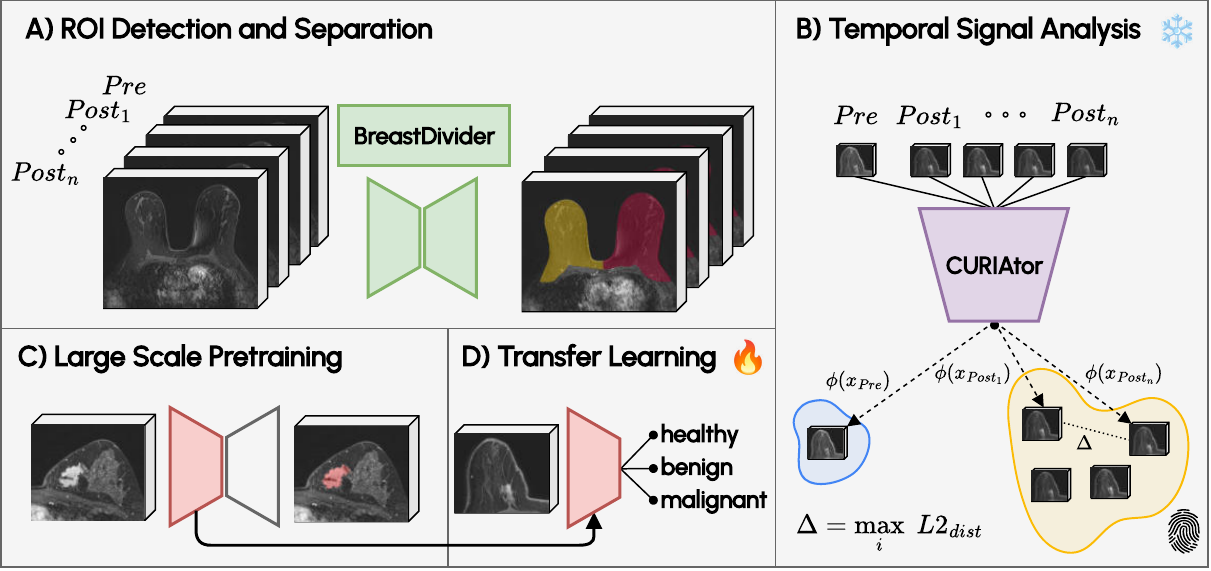}
 \caption{Overview of the proposed foundation model-driven framework for breast cancer classification from DCE-MRI. The framework includes A) the \textit{BreastDivider} model for ROI detection and breast separation, B) adaptive temporal signal analysis with the \textit{CURIAtor} module selects the most informative features, C) large-scale anatomically informed pretraining, and D) transfer learning for final lesion classification.}
 \label{fig:architecture}
\end{figure}

Existing approaches attempt to cope with the temporal and volumetric complexity of DCE-MRI through aggressive input simplifications. A common strategy is to predefine a fixed subset of temporal phases and applying this same configuration to all cases irrespective of protocol differences or lesion specific enhancement patterns \cite{khaled2022u,alanazi2026domain,satake2025predictive,yin2023combined}. While computationally convenient, such fixed phase selection implicitly assumes consistent temporal sampling across centers and does not account for variations in the number of acquired phases or irregular inter scan intervals. Other methods reduce dimensionality by applying maximum intensity projections (MIP) to individual 3D volumes or subtraction volumes independently, thereby collapsing the volumetric spatial information into multi-channel 2D representations \cite{jing2022using,antropova2018use}. Although this strategy reduces memory requirements and simplifies modeling, it introduces irreversible information loss by compressing depth and contrast uptake dynamics into a single intensity representation. In particular, subtle enhancement differences may be suppressed, while surrounding regions that enhance strongly at any time point can dominate the projection, potentially obscuring small lesions and diagnostically relevant kinetic patterns. These limitations motivate the need for a method that explicitly structures spatial and temporal invariance rather than relying on static dimensionality reduction, as existing approaches remain vulnerable to protocol-specific biases that contribute to degraded cross-center generalization in multi-site settings \cite{alanazi2026domain}.\\
\indent To overcome these challenges, we introduce a framework that dynamically adapts to the latent characteristics of any given protocol for robust breast cancer classification from DCE-MRI. Rather than simplifying the input through static phase selection or projection, our approach explicitly organizes data modeling into successive levels of abstraction that address the three dominant sources of variability in clinical DCE-MRI.\\
\textbf{First}, spatial variability is addressed through anatomically informed localization that reduces confounding background uptake and compensates for misalignment caused by deformable soft tissue. \\
\textbf{Second}, temporal variability is addressed through adaptive relational modeling of enhancement dynamics. The Curia radiology foundation model (FM) \cite{dancette2025curia} first embeds each temporal volume into a high level feature space, after which the representation constructs relationships across phases and dynamically selects the most informative temporal features. This enables the model to accommodate differences in phase number, timing, and sampling across centers without relying on a fixed acquisition scheme or predefined phase selection. \\
\textbf{Third}, to compensate for the weak supervisory signal inherent to breast-level classification, we leverage large-scale anatomically grounded pretraining, as pure classification, unlike segmentation, provides no explicit spatial guidance toward lesion regions. Pretraining on anatomically structured tasks therefore shapes feature representations before finetuning, improving sensitivity to small pathological patterns despite the absence of pixel level supervision~\cite{wald2025enhancing}. \\
To the best of our knowledge, we are the first to introduce a temporally adaptive component rather than relying on a static handling of dynamic acquisitions. Furthermore, we are the first to integrate spatial alignment, temporal alignment, and supervised pretraining into a unified three stage hierarchy, forming a single coherent framework for robust breast cancer classification.

\section{Methods}

Our method follows a multi-level representation hierarchy that progressively abstracts anatomical, temporal, and task-specific information from heterogeneous DCE-MRI data. A hierarchical overview of the proposed framework is shown in Fig. \ref{fig:architecture}.

\subsection{BreastDivider Model}

First, we propose BreastDivider, a nnU-Net–based \cite{isensee2021nnu} 3D segmentation model trained to localize and separate left and right breast regions in clinical MRI volumes. It is trained on a curated large-scale dataset comprising 17,956 scans aggregated from multiple public cohorts spanning more than 55 centers \linebreak \cite{duke,ispy1,ispy2,mamamia,ambl,ea1141,odelia,rokuss2025divide}. The dataset covers a broad spectrum of acquisition protocols, including T1, contrast enhanced T1, T2, FLAIR, and diffusion weighted imaging, thereby capturing substantial inter center and inter sequence variability. The dataset was constructed using an iterative active learning strategy, starting from the Duke cohort \cite{duke} and progressively expanding. In successive training rounds, only high quality and informative examples were retained for subsequent optimization. A sample was deemed reliable when all five cross-validation folds achieved an agreement exceeding 0.95 Dice overlap. Training on this diverse and anatomically curated resource enables BreastDivider to achieve highly accurate localization performance (Dice: $99.08 \pm 0.48$) and to serve as a robust preprocessing component for downstream tasks.

\subsection{CURIAtor Module}

DCE-MRI acquisitions are inherently redundant, with diagnostically relevant information concentrated in a few physiologically distinct phases, particularly around the early post contrast period \cite{kuhl2024abbreviated,kuhl2014abbreviated}. Clinical evidence from abbreviated and ultrafast breast MRI protocols shows that combining pre contrast imaging, the first post contrast phase, and a delayed phase capturing complementary enhancement can achieve performance close to full dynamic acquisitions \cite{cao2024optimizing}. Motivated by this observation, we design the \textit{CURIAtor} module to operationalize this principle. Given the established importance of the pre contrast and first post contrast volumes, the module adaptively identifies a third phase that provides the most complementary temporal signal for the specific acquisition. Concretely, we select the most \textbf{dissimilar} post contrast phase. Drawing on principles from diversity-based active learning, samples that are far apart in feature space tend to provide the most new information relative to already observed data \cite{ash2019deep}.

\begin{figure}[ht]
 \centering
 \includegraphics[width=\linewidth]{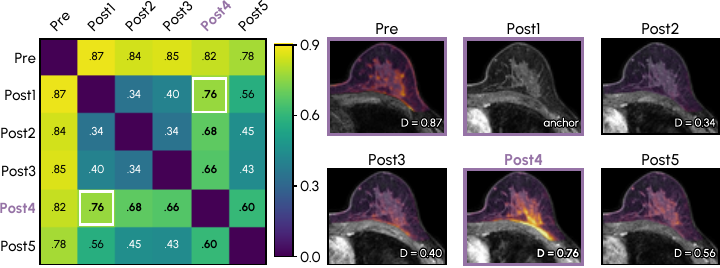}
 \caption{Pairwise distance matrix $\bm{D}$ computed by the \textit{CURIAtor} module for UMCU 0178187682 Right (ODELIA).
 Each entry encodes the volume-level similarity between two temporal acquisitions within the anatomically constrained ROI; lower values indicate higher similarity.
 Post4 is farthest from Post1 and from every other post-contrast phase, and is selected (purple) alongside Pre and Post1.
 Overlay: $|V^n - V^{\mathrm{Post}_1}|$ within the ROI on one axial slice.}
 \label{fig:latent}
\end{figure}

Formally, let $\mathcal{V} = \{V^n\}_{n=1}^{N}$ be the set of $N$ temporal volumes of a single subject, where each volume $V^n \in \mathbb{R}^{D \times H \times W}$ represents one acquisition. The BreastDivider mask $M \in \{0,1\}^{D \times H \times W}$ constrains all subsequent processing to the anatomically relevant region. All volumes are first cropped to the minimal three-dimensional bounding box enclosing the mask foreground, eliminating the large surrounding background that would otherwise contribute noise and increase computational cost. Each normalized slice $V^n_d$ is independently processed by the pretrained Curia radiology FM \cite{dancette2025curia}, yielding slice-level features $X^n_d \in \mathbb{R}^{P \times C}$, where $P$ is the number of spatial tokens and $C$ is the embedding dimension. To improve robustness to local slice-to-slice misalignment, tokens are aggregated across slices using average pooling with a kernel size of $8$, resulting in a volume-level embedding $E^n \in \mathbb{R}^{M \times C}$, where $M = D_r \cdot P$ and $D_r$ denotes the pooled depth. Anatomical relevance is further enforced by projecting the ROI mask onto the token grid, yielding fractional overlap weights $w_{d,p} \in [0,1]$ that down-weight contributions from background tokens in all subsequent distance computations. The pairwise dissimilarity between volumes $i$ and $j$ is computed as a masked, spatially weighted distance. For each slice $d$, the token-level squared distances are aggregated under the ROI weights as:

\begin{equation}
  D^{ij}_{d_r} = \sqrt{\frac{\sum_p w_{{d_r},p}\;\delta^{ij}_{{d_r},p}}{\sum_p w_{{d_r},p}}},
  \label{eq:slice_dist}
\end{equation}

where $\delta^{ij}_{{d_r},p} = \frac{1}{C}\sum_c (E^i_{m,c} - E^j_{m,c})^2$ is the normalized squared Euclidean distance at token position $m = ({d_r}, p)$. Slice-level distances are averaged over the cropped depth to yield a scalar volume-level dissimilarity $D_{ij} = \frac{1}{D_r} \sum_{d_r} D^{ij}_{d_r}$. Collecting all pairwise values
yields a symmetric distance matrix $\mathbf{D} \in \mathbb{R}^{N \times N}$, which defines a similarity manifold over the full acquisition sequence. Taking the first post-contrast volume $V^{\mathrm{Post}_1}$ as a fixed anchor \textit{CURIAtor} selects the most informative remaining phase by maximizing its dissimilarity to $V^{\mathrm{Post}_1}$: $n^* = \operatorname*{arg\,max}_{n \geq 3}\; D_{2n}$. The resulting triplet $\{V^{\mathrm{Pre}},\, V^{\mathrm{Post}_1},\, V^{n^*}\}$ constitutes a protocol-invariant temporal representation of the acquisition. Because the selection is driven entirely by the learned embedding geometry of Curia, the module requires no explicit phase labels, no predefined kinetic descriptors, and no assumptions about sampling regularity. The representation is therefore robust to variable phase counts, irregular inter-scan intervals, and scanner-specific acquisition differences across clinical centers. Sample volume level similarities are visualized in Fig.~\ref{fig:latent}.

\subsection{Transfer Learning Framework}

Since self supervised learning often struggles to capture small pathologies and subtle abnormal regions \cite{wald2025revisiting}, we adopt an anatomically informed transfer learning strategy to initialize and adapt our classification models across heterogeneous breast MRI datasets. We first perform breast tumor segmentation pretraining on the MAMA-MIA cohort by splitting bilateral examinations into individual breast regions, resulting in more than 17,000 three dimensional breast volumes derived from 3,012 breast instances acquired across multiple imaging sequences. All inputs are treated as single channel images, enabling the encoder to learn sequence agnostic representations and to capture the full range of intensity and contrast variability present across acquisition protocols. This is achieved by training nnU-Net for 2000 epochs using the \textit{ResEncL} configuration \cite{isensee2024nnu}. The pretrained encoder is then used as a universal initialization for the downstream classification task.

\section{Experiments and Results}

We finetune the classification model from the pretrained weights for 100 epochs using SGD with cross-entropy loss and weight decay of $3 \times 10^{-5}$. The learning rate follows a polynomial decay schedule with a 10-epoch linear warmup and a peak value of $1 \times 10^{-3}$. We evaluate our framework with focus on cross center generalization for the classification of healthy (H), benign (B), and malignant (M) breast tumors, as summarized in Table~\ref{tab:comparison}. We perform five fold cross validation on the ODELIA dataset, where each fold corresponds to a distinct clinical center to enforce strict separation across acquisition sites, scanners, and imaging protocols and thus provide a realistic assessment of distribution shift, and additionally evaluate on the held out AMBL dataset. Subgroup analyses are precluded due to the lack of harmonized cohort metadata. Both datasets are dominated by healthy cases (ODELIA: 683 H, 137 B, 202 M; and AMBL: 190 H, 91 B, 111 M), with AMBL comparatively more balanced across the three classes. Further acquisition details, such as scanner parameters, can be found in the respective original publications~\cite{odelia,ambl}. We benchmark our approach against finetuned FMs, namely Merlin \cite{blankemeier2024merlin}, VISTA3D \cite{he2025vista3d}, CT-FM \cite{pai2025vision}, and Curia \cite{dancette2025curia}. We leave native 4D model architectures as future work. All non-CURIA baselines were fine-tuned end-to-end with input resolution and CURIAtor-selected inputs matched to our method; We used AdamW with a cosine schedule (100 epochs, batch 32, 50 steps/epoch). For CURIA: Frozen 2D embeddings, slice-wise mean aggregation and linear probe. Furthermore, we compare against existing approaches based on MIP. Finally, we conduct ablation studies to assess the contributions of the transfer learning strategy, the ROI prediction, and the \textit{CURIAtor} module by comparing adaptive temporal selection against static input configurations.

\noindent\textbf{Best cross-institutional generalization}. Our framework consistently outperforms radiology FMs across all centers, suggesting that general-purpose models, while effective for generic feature extraction, are less suited for finetuning on tasks with localized targets. Moreover, the improvements over MIP‑based approaches underscore the limitations of aggressively collapsing depth dimensions.

\begin{table}[t]
\centering
\caption{Cross-center performance comparison on the ODELIA\cite{odelia} and AMBL\cite{ambl} datasets. Results are reported in terms of AUROC and balanced accuracy (BA) for each clinical center and on AMBL, as well as their mean. Bootstrap 95\% CIs in brackets.}
\setlength{\tabcolsep}{3pt}
\renewcommand{\arraystretch}{0.9}
\setlength{\aboverulesep}{0.8pt}
\setlength{\belowrulesep}{0.8pt}
\newcommand{\rawstyle}[1]{#1}

\newcommand{\valcia}[4][\rawstyle]{%
  \ifdim #2 pt < \minvale pt
    \cellcolor{goodred}%
  \else\ifdim #2 pt > \maxvale pt
    \cellcolor{goodgreen}%
  \else\ifdim #2 pt < \midvale pt
    \pgfmathsetmacro{\PercentLow}{200.0*(#2-\minvale)/(\maxvale-\minvale)}%
    \xdef\PercentLow{\PercentLow}%
    \cellcolor{goodyellow!\PercentLow!goodred}%
  \else
    \pgfmathsetmacro{\PercentHigh}{200.0*(#2-\midvale)/(\maxvale-\minvale)}%
    \xdef\PercentHigh{\PercentHigh}%
    \cellcolor{goodgreen!\PercentHigh!goodyellow}%
  \fi\fi\fi
  \begin{tabular}[c]{@{}c@{}}#1{#2}\\[-3.5pt]{\scriptsize#1{[#3,\,#4]}}\end{tabular}%
}

\newcommand{\valcib}[4][\rawstyle]{%
  \ifdim #2 pt < \minval pt
    \cellcolor{goodred}%
  \else\ifdim #2 pt > \maxval pt
    \cellcolor{goodgreen}%
  \else\ifdim #2 pt < \midval pt
    \pgfmathsetmacro{\PercentLow}{200.0*(#2-\minval)/(\maxval-\minval)}%
    \xdef\PercentLow{\PercentLow}%
    \cellcolor{goodyellow!\PercentLow!goodred}%
  \else
    \pgfmathsetmacro{\PercentHigh}{200.0*(#2-\midval)/(\maxval-\minval)}%
    \xdef\PercentHigh{\PercentHigh}%
    \cellcolor{goodgreen!\PercentHigh!goodyellow}%
  \fi\fi\fi
  \begin{tabular}[c]{@{}c@{}}#1{#2}\\[-3.5pt]{\scriptsize#1{[#3,\,#4]}}\end{tabular}%
}
\adjustbox{width=\textwidth}{
\begin{tabular}{l|ccccc|c|c}
\toprule
Dataset & CAM & MHA & RUMC & UKA & UMCU & AMBL & Mean \\
\midrule
\multicolumn{8}{c}{\textbf{AUROC [\%] $\uparrow$}} \\
\midrule
Curia\cite{dancette2025curia}
& \valcia{53.5}{48.5}{58.9} & \valcia{50.5}{42.0}{60.3} & \valcia{37.6}{16.6}{59.4}
& \valcia{51.4}{44.5}{58.5} & \valcia{53.7}{47.1}{59.9} & \valcia{52.2}{47.7}{56.8}
& 49.8 \\
VISTA3D\cite{he2025vista3d}
& \valcia{54.1}{48.8}{59.6} & \valcia{62.0}{53.3}{70.1} & \valcia{57.1}{33.0}{82.0}
& \valcia{49.5}{43.1}{55.7} & \valcia{61.6}{55.1}{68.0} & \valcia{51.5}{47.6}{55.7}
& 56.0 \\
CT-FM\cite{pai2025vision}
& \valcia{52.0}{47.2}{57.5} & \valcia{49.1}{40.5}{57.3} & \valcia{35.9}{25.6}{46.9}
& \valcia{48.6}{42.0}{55.6} & \valcia{61.4}{54.6}{67.6} & \valcia{55.4}{51.3}{59.7}
& 50.4 \\
Merlin \cite{blankemeier2024merlin}
& \valcia{64.0}{59.1}{68.6} & \valcia{59.0}{49.7}{68.9} & \underline{\valcia{67.7}{52.6}{83.0}}
& \valcia{48.4}{42.1}{55.2} & \valcia{55.2}{48.3}{61.5} & \underline{\valcia{56.0}{52.6}{60.1}}
& 58.4 \\
\midrule
MIP\cite{jing2022using,antropova2018use}
& \underline{\valcia{72.4}{65.8}{79.4}} & \valcia{67.6}{60.5}{74.9} & \valcia{67.7}{60.6}{74.7}
& \underline{\valcia{70.3}{62.7}{77.5}} & \textbf{\valcia{76.6}{69.4}{83.0}} & \valcia{49.8}{47.9}{52.0}
& 67.4 \\
\midrule
\textbf{Ours}
& \valcia[\textbf]{82.7}{78.1}{87.5} & \valcia[\textbf]{79.0}{70.4}{86.5}
& \valcia[\textbf]{72.8}{60.4}{86.0} & \valcia[\textbf]{73.5}{67.2}{79.2}
& \valcia[\underline]{71.7}{66.6}{76.7} & \valcia[\textbf]{71.1}{69.3}{72.7}
& \textbf{75.1} \\
\midrule
\multicolumn{8}{c}{\textbf{Balanced Accuracy [\%] $\uparrow$}} \\
\midrule
Curia \cite{dancette2025curia}
& \valcib{33.6}{32.2}{35.2} & \valcib{33.0}{32.3}{33.3} & \valcib{22.6}{3.10}{42.4}
& \valcib{36.2}{29.8}{43.1} & \valcib{36.1}{30.1}{41.7} & \valcib{33.2}{32.8}{33.3}
& 32.4 \\
VISTA3D\cite{he2025vista3d}
& \valcib{33.2}{33.0}{33.3} & \valcib{43.2}{37.9}{48.1} & \valcib{46.1}{26.7}{64.8}
& \valcib{34.2}{26.8}{41.8} & \valcib{46.7}{40.9}{52.4} & \valcib{33.8}{33.0}{34.8}
& 39.5 \\
CT-FM\cite{pai2025vision}
& \valcib{36.6}{31.8}{41.6} & \valcib{37.4}{31.4}{43.3} & \valcib{32.4}{30.1}{50.0}
& \valcib{33.6}{25.8}{41.7} & \valcib{42.3}{36.8}{47.6} & \valcib{37.3}{34.7}{40.3}
& 36.6 \\
Merlin\cite{blankemeier2024merlin}
& \valcib{48.0}{44.2}{52.0} & \valcib{44.5}{39.0}{49.7} & \valcib{38.7}{30.5}{65.2}
& \valcib{30.9}{24.3}{37.5} & \valcib{43.2}{37.9}{48.1} & \valcib{39.9}{37.0}{42.9}
& 40.9 \\
\midrule
MIP\cite{jing2022using,antropova2018use}
& \underline{\valcib{63.3}{58.0}{68.7}} & \underline{\valcib{61.5}{56.4}{66.7}} & \underline{\valcib{56.7}{52.7}{61.4}}
& \underline{\valcib{61.3}{56.3}{66.6}} & \textbf{\valcib{69.3}{64.0}{74.7}} & \underline{\valcib{50.1}{49.5}{50.8}}
& 60.4 \\
\midrule
\textbf{Ours}
& \valcib[\textbf]{72.5}{69.7}{75.1} & \valcib[\textbf]{64.5}{59.4}{70.5}
& \valcib[\textbf]{62.2}{52.0}{71.2} & \valcib[\textbf]{61.7}{57.5}{65.8}
& \valcib[\underline]{63.2}{58.0}{68.9} & \valcib[\textbf]{62.5}{61.1}{63.9}
& \textbf{64.4} \\
\bottomrule
\end{tabular}
}
\label{tab:comparison}
\end{table}

\noindent\textbf{Superior adaption to protocol variability}. Table~\ref{tab:ablations_config} reports ablations isolating each component's contribution. Removing pretraining clearly degrades performance, confirming the value of transfer learning for robust representation initialization. Replacing the BreastDivider ROI prediction with whole-volume processing also sharply reduces performance, underscoring the benefit of anatomically constrained feature extraction. Finally, the \textit{CURIAtor} selection consistently outperforms static sequence configurations, showing that dynamic temporal weighting provides significant discriminative signal.

\begin{table}[ht]
\centering
\setlength{\tabcolsep}{3pt}
\caption{Ablation study evaluating the contribution of each component of the proposed framework. Performance is reported in terms of mean AUROC and balanced accuracy (BA) across centers. $\Delta\%$ denotes the relative performance change with respect to the final framework using \textit{CURIAtor}.}
\label{tab:ablations_config}
\adjustbox{width=\textwidth}{
\begin{tabular}{lccc cc}
\toprule
 & \makecell{\textbf{ROI}} & \makecell{\textbf{\textit{CURIAtor}}} & \makecell{\textbf{Pretraining}} &
 \makecell{\textbf{Mean}\\\textbf{AUROC}} & \makecell{\textbf{Mean}\\\textbf{BA}} \\
\midrule
\multicolumn{6}{l}{\textbf{Ablation [$\Delta\%$]}} \\
\midrule
No Pretraining & \checkmark & \checkmark & -- & \cellcolor{red!25} -19.19 & \cellcolor{red!25} -15.61 \\
Whole Volume & -- & \checkmark & \checkmark & \cellcolor{red!25} -19.89 & \cellcolor{red!25} -15.00 \\
\midrule
\multicolumn{6}{l}{\textbf{Input Sequences [$\Delta\%$]}} \\
\midrule
$Pre$ + $Post_{n/2}$ + $Post_n$ + $T2$  & \checkmark & -- & \checkmark & \cellcolor{red!25} -11.33 & \cellcolor{red!25} -3.16 \\
$Pre$ + $Post_{n/2}$ + $Post_n$ & \checkmark & -- & \checkmark & \cellcolor{red!25} -11.32 & \cellcolor{red!25} -3.15 \\
$Pre$ + $Post_1$ + $Post_n$ & \checkmark & -- & \checkmark & \cellcolor{red!25} -11.65 & \cellcolor{red!25} -3.22 \\
$Pre$ + $Post_1$ + $Post_2$ & \checkmark & -- & \checkmark & \cellcolor{red!25} -4.86 & \cellcolor{red!25} -1.66 \\
\midrule
\textbf{Ours [\%]}
& \checkmark & \checkmark & \checkmark
& \cellcolor{green!25}\textbf{75.13}
& \cellcolor{green!25}\textbf{64.42} \\
\bottomrule
\end{tabular}
}
\end{table}

\section{Discussion and Conclusion}

In this work, we propose a comprehensive framework for robust breast cancer classification from DCE-MRI. Our approach addresses three central challenges of clinical DCE-MRI analysis: high dimensional 4D inputs, small lesion extent relative to background anatomy, and substantial inter center protocol variability. To the best of our knowledge, we are the first to introduce a native 4D temporally adaptive component rather than relying on static phase selection or collapsing volumes using MIP. We further integrate spatial alignment, temporal alignment, and supervised lesion pretraining into a unified three stage hierarchy, forming a single coherent architecture to address the outlined challenges. Extensive cross center evaluation on ODELIA and AMBL demonstrates consistent and substantial improvements over finetuned radiology foundation models and MIP based baselines under identical training conditions. Our method achieves a mean AUROC of 75.13 and a mean BA of 64.42 across centers, outperforming the strongest MIP baseline by nearly 8 AUROC points and 4 BA points. Replacing \textit{CURIAtor} with static phase combinations reduces performance by up to 11.65 AUROC points and over 3 BA points. The observed performance gains indicate that robust generalization under real world distribution shift stems not from increasing model capacity or self supervised pretraining scale alone, but from explicitly structured, anatomy aware and temporally adaptive representation design. In addition, our method achieved first place in the MICCAI ODELIA Breast MRI Challenge 2025~\cite{hamm2025meisenmeister}, further demonstrating its effectiveness for robust clinical breast cancer classification.

\begin{credits}
\subsubsection{\ackname} This work was supported by the Helmholtz Association under the joint research program “HIDSS4Health – Helmholtz Information and Data Science School for Health” and under the Helmholtz Foundation Model Initiative (HFMI), project “The Human Radiome Project” (THRP). This work was partially funded by Helmholtz Imaging, a platform of the Helmholtz Information \& Data Science Incubator, by “NUM 2.0“ (FKZ: 01KX2121), by “NUM 3.0” (FKZ: 01KX2524), by the Deutsche Forschungsgemeinschaft (DFG, German Research Foundation) under project number 402688427, and by the Helmholtz AI project “Effective Privacy-Preserving Adaptation of Foundation Models for Medical Tasks” (PAFMIM; ZT-I-PF-5-227). Maximilian Rokuss was supported by the Google PhD Fellowship Program.

\subsubsection{\discintname}
Maximilian Rokuss was supported by the Google PhD Fellowship Program. Other than that the authors have no competing interests to declare that are relevant to the content of this article.
\end{credits}

\bibliographystyle{splncs04}
\bibliography{Paper-3872}

\end{document}